\documentclass[journal]{IEEEtran}
\IEEEoverridecommandlockouts
\usepackage[T1]{fontenc}
\usepackage{cite}

\usepackage{amsmath}
\usepackage{times} 
\usepackage{amssymb}  
\usepackage{graphicx}
\usepackage{subfigure}
\usepackage{bm}
\usepackage{physics}
\usepackage{threeparttable}
\usepackage{gensymb}
\usepackage{fancyhdr}
\usepackage{booktabs}
\usepackage{float}
\usepackage{bigstrut}
\usepackage{multirow}
\usepackage{balance}
\usepackage{url}
\usepackage{bm}
\usepackage{booktabs}
\usepackage{float}
\usepackage{bigstrut}
\usepackage{caption}
\usepackage{amsopn}
\usepackage{epstopdf} 
\usepackage[vlined,ruled]{algorithm2e}
\usepackage[table]{xcolor}
\usepackage[mathlines,switch]{lineno}
\usepackage{placeins}
\usepackage{amsbsy}
\usepackage{amsthm}

\usepackage{booktabs,makecell,multirow}
\usepackage{url}
\usepackage{xcolor}

\usepackage{fancyhdr}

\fancypagestyle{plain}{
  \fancyhf{} 
  \fancyhead[L]{} 
  \fancyhead[C]{} 
  \fancyhead[R]{}
  \fancyfoot[L]{\thepage}

}

\def\BibTeX{{\rm B\kern-.05em{\sc i\kern-.025em b}\kern-.08em
    T\kern-.1667em\lower.7ex\hbox{E}\kern-.125emX}}

\begin{document}

\title{Intelligent Collective Escape of Swarm Robots Based on a Novel Fish-inspired Self-adaptive Approach with Neurodynamic Models}
\author{Junfei Li,~\IEEEmembership{ Member,~IEEE,} and Simon X. Yang,~\IEEEmembership{Senior Member,~IEEE }

\thanks{This work was supported by the Natural Sciences and Engineering Research Council (NSERC) of Canada. (Corresponding author:
Simon X. Yang.)}
\thanks{The authors are  with  the  Advanced  Robotics  and  Intelligent  Systems  Laboratory, School  of Engineering,  University  of Guelph,  Guelph,  ON N1G2W1, Canada. Emails: {\{jli64;syang\}@uoguelph.ca}}
}

\maketitle
\thispagestyle{fancy}
\begin{abstract}
Fish schools present high-efficiency group behaviors through simple individual interactions to collective migration and dynamic escape from the predator. The school behavior of fish is usually a good inspiration to design control architecture for swarm robots. In this paper, a novel fish-inspired self-adaptive approach is proposed for collective escape for the swarm robots. In addition, a bio-inspired neural network (BINN) is introduced to generate collision-free escape robot trajectories through the combination of attractive and repulsive forces. Furthermore, to cope with dynamic environments, a neurodynamics-based self-adaptive mechanism is proposed to improve the self-adaptive performance of the swarm robots in the changing environment. Similar to fish escape maneuvers, simulation and experimental results show that the swarm robots are capable of collectively leaving away from the threats. Several comparison studies demonstrated that the proposed approach can significantly improve the effectiveness and efficiency of system performance, and the flexibility and robustness in complex environments.

\end{abstract}

\begin{IEEEkeywords}
Self-adaptive motion, escape behaviors, swarm robots, neural networks, and bio-inspired algorithms.
\end{IEEEkeywords}

\vspace{-0.15cm}
\section{Introduction}
The self-organizing collective behaviors are wildly observed in nature, where a large number of group animals are able to accomplish magnificent cooperative behaviors only depending on relatively simple interactions \cite{ioannou2017swarm}. The collective behaviors enable animals as a whole to be greater than the sum of individuals.
Recently, many types of research have involved exploiting the understanding of mechanisms to achieve the collective behaviors of swarm robots \cite{zhang2023distributed,yu2020intelligent,roy2020geometric}.

The escape of multiple robots is a classic and prevalent issue in the research field of robotics. Traditional approaches considered robot escape as a pursuit-evasion game, in which the robot tries to avoid being captured by the threat \cite{tian2021distributed,selvakumar2019feedback,zha2016construction}. However, traditional approaches exhibit a lack of environmental adaptability and fall short in facilitating inter-robot collaboration. 
Fish are known to have the ability to navigate and respond effectively to dynamic environments through the use of simple mechanisms. Through cooperation and limited implicit communication, fish schools are able to accomplish complex tasks that would be beyond the capabilities of an individual fish \cite{doran2022fish}. 
Methods about how to achieve collective escape behaviors in robotics are currently an active focus of research. The fountain maneuver model is able to reproduce many collective features in fish schools \cite{ishiwaka2021foids}. Thus, the fountain maneuver model was copied into the robotic system, providing a solution to control robots with distributed control architecture.  Cioarga \textit{et al.} \cite{cioarga2010evaluation} implemented collision-free fountain maneuvers, and variations of flash expansion based on mobile robots. Berlinger \textit{et al.} \cite{berlinger2021self} used the fountain maneuver model for the underwater robotic platform while keeping a constantly visible angle to the predator. 
The artificial virtual forces from nearby neighbors based on relative distance are another method of achieving the collective behaviors of swarm robots.  Berlinger \textit{et al.} \cite{berlinger2021implicit} used a potential field-based model to mimic collective behaviors for underwater robots in  3D environments. Novák \textit{et al.} \cite{novak2021fast} proposed an animal-inspired fast escape method that allows swarm robots to avoid dynamic obstacles. Min and Wang \cite{min2011design} proposed a fish-inspired escape algorithm based on Newton-Euler dynamics equations, which is able to escape the predator rapidly and avoid collision with obstacles.  
As mentioned in the literature reviews, the limitations of most existing studies are summarized as follows:

\begin{enumerate}
\item  Many studies only considered the collision-free or static obstacle environment 
\item  Many studies required that the robot has full knowledge about the environment, including the position of threat, obstacle, and other robots. 
\item  Many studies have become computationally expensive when considering the large size of robots. 
\end{enumerate}

The purpose of this paper is to infuse swarm robots with ``fish-like” intelligence and properties that will enable safe navigation and efficient cooperation among the autonomous robots, and successful completion of escape tasks in changing environments.
The swarm robots are considered as a group of simple robots that are autonomous and only have limited neighbor information. It is important to note that the proposed collective escape is fish-inspired, but the core operation departs from the real fish through the introduction of the neurodynamics model. In real-world fish schools, the individual fish can detect the water movement to perceive neighbors through their lateral lines \cite{ioannou2017swarm}. In this paper, the neurodynamics model is incorporated to improve self-adaptive performance in dynamic environments. In addition, a bio-inspired neural network (BINN) is introduced to generate collision-free escape trajectories and virtual forces. 
The proposed fish-inspired self-adaptive approach to the collective escape of the swarm robots has been shown to have several advantages over existing escape methods, including faster operation, higher efficiency, and improved reliability. 
Simulation and experimental results have demonstrated the effectiveness of the proposed approach for ensuring safe escape and efficient self-adaptive cooperation among autonomous robots in changing environments. 
The main contributions of this paper are summarized as follows:
\begin{itemize}
		\item A novel collective escape is proposed to swarm robots in changing environments. The proposed approach is inspired by the group behavior of fish with only local sensing ability. 
  
            \item A novel collision-free virtual forces approach is proposed to guide swarm robots based on the BINN. During the escape process, there are no learning procedures for the movement of robots.

        \item A novel neurodynamics-based self-adaptive mechanism is proposed to consider the effects of obstacles, which enables swarm robots to dynamically adjust their parameters in complex environments.
	     

\end{itemize}

This paper is organized in the following manner. Section \ref{sec:Problem} gives the problem statement. Section \ref{sec:approach} describes the proposed approaches. Section \ref{sec:simulation} shows the simulation and  comparison results. Section \ref{sec:experiments} provides a real robot experiment.  Section \ref{sec:discussion} analyses the characteristic of the neurodynamics model.
In Section \ref{sec:conclusion}, the result is briefly summarized.



\section{Problem Statement}
\label{sec:Problem}
For a swarm of $m$ robots, their time-varying location at time instant $t$ in the 2D Cartesian workspace $W$, can be uniquely determined by the spatial position $\mathbf{p}_{e}= (x_e,y_e)$, $e=1,\ldots,m$. The maximum speed of the $e$-th robot is denoted by $V_{max}>0$. Suppose each robot is considered as an omnidirectional robot, which can change the moving direction without delay. The next location of the $e$-th robot at time instant $t+1$ can be given as 
\begin{equation}
\left(x_{e}\right)_{t+1}=\left(x_{e}\right)_{t}+v_{k} \Delta t \cos \left(\theta_{e}\right)_{t}
\end{equation}
\begin{equation}
\left(y_{e}\right)_{t+1}=\left(y_{e}\right)_{t}+v_{k} \Delta t \sin \left(\theta_{e}\right)_{t}
\end{equation}
where the $v_k<V_{max}$ is the current speed of the robot; $\theta_{e}$ is the moving direction of the robot; and $\Delta t $ is the unit time interval. In addition, there is a sequence of static and moving obstacles in $W$. Let $\mathcal{O}$ be an obstacle scenario. 
The time-varying  collision-free area pertaining to $\mathcal{O}$ can be defined as $\mathcal{O}_{\text {free }}^t=\left\{(x,y) \in \mathbb{R}^2: \Gamma >1\right\}$, where 
\begin{equation}
\Gamma=\frac{\left(x-x_o^t\right)^{2}+\left(y-y_o^t\right)^{2}}{\lambda_o}
\end{equation}
where $(x_o^t,y_o^t)$ is the center of the obstacle which is updated with respect to a constant moving speed  $v_o$ and $\lambda_o$ is the size of the obstacle. The regions meeting $\Gamma =1$, $\Gamma >1$, or $\Gamma <1$ denote the surface, exterior, or interior of the obstacle, respectively. 
The robot is equipped with sensors with $360\degree$ visual capability and a detection range $R_s$ to recognize the position of the threat, obstacle and other robots. The input for $e$-th robot can be represented as
\begin{equation}
\mathcal{I}_{{e }}=\left\{\mathbf{p}_{d}, M_d, \mathcal{O}, \mathcal{T}\right\}
\end{equation}
where $\mathbf{p}_{d}$ is the position of the $d$-th robot within the detection range $R_s$; $M_d$ is the current mode state of the $d$-th robot and $\mathcal{T}$ is the position of the threat within the detection range $R_s$.
Assume that there is no inter-robot communication between individual robots. Therefore, the robots cannot share neighboring robots information $\mathbf{p}_{d}$ and $M_d$, or environmental information $\mathcal{O}$ and $\mathcal{T}$ with other robots. The input of the robot is solely dependent on its own observations.
In addition, the robots is required self-adaptive to changing environments and maintain the desired distance $R_d$ from their neighbors during the escape process.

Therefore, the fish-inspired escape studied in this paper can be described as: for a group of $m$ robots and given the initial positions of robots $\mathbf{p}_e(0)$ with $e=1,\ldots,m$. Since the $i$-th robot detects a threat, the $i$-th robot begins to escape and generate a collision-free trajectory, that is, $\mathbf{P}_{\text {Escape}} \in \mathcal{O}_{\text {free }}^t$, until achieving a safe distance $d_s$. Other robots generated collision-free trajectories, that is, $\mathbf{P}_{\text {Follow}} \in \mathcal{O}_{\text {free }}^t$, to follow the $i$-th robot with the desired distance $R_d$ to neighboring robots. During the escape process, the swarm robots are required to be self-adaptive to adapt to changing environments. 


\section{Proposed Approach}
\label{sec:approach}
In this section, a fish-inspired system for robot escape behavior and a neural network structure are introduced to generate collision-free virtual forces. In addition, a self-adaptive mechanism based on the dynamic landscape of neural activity is proposed.

\subsection{Fish-inspired Behavioral Modeling and Organization}
\begin{figure}[t]
\centering
\includegraphics[width=3.4in]{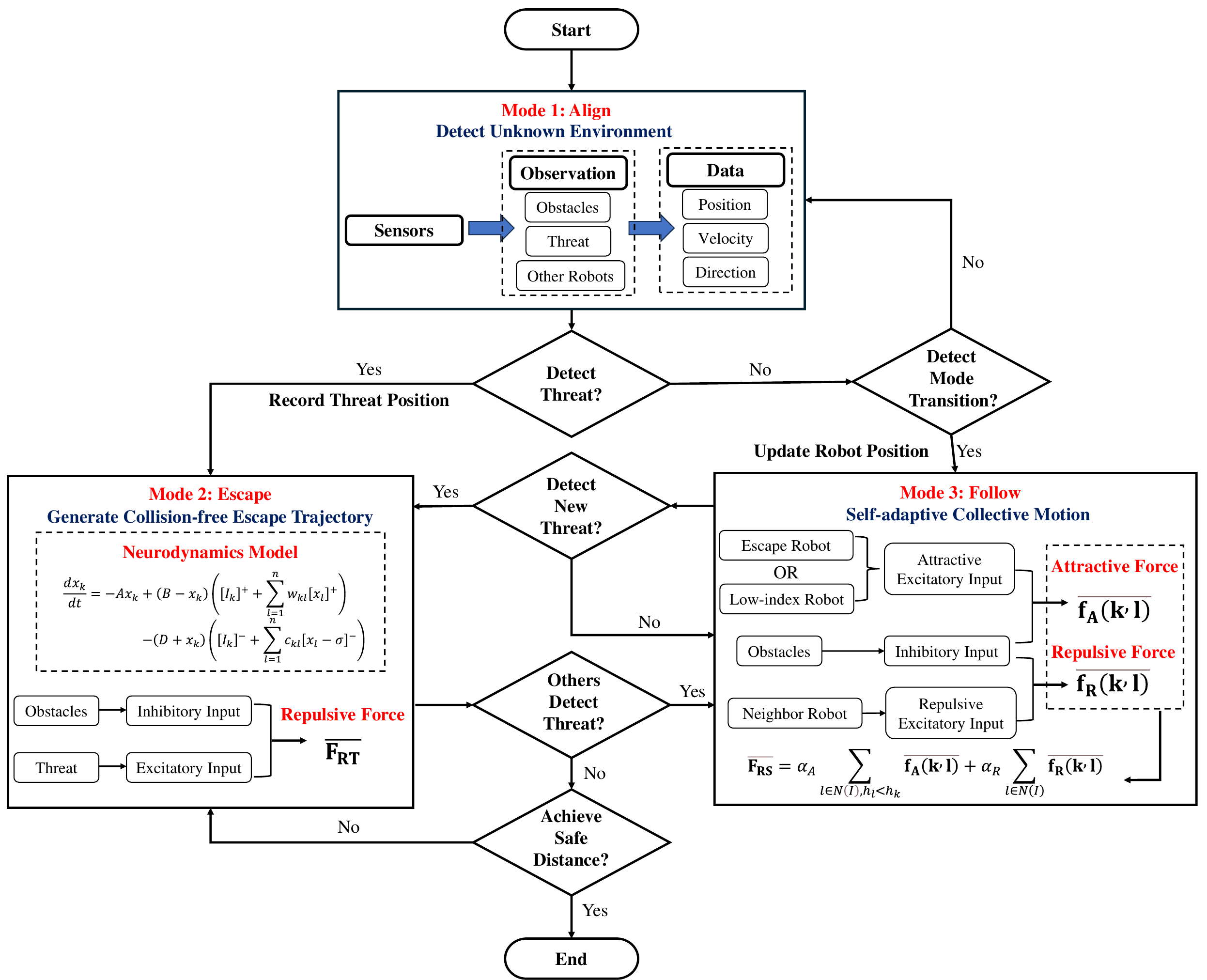}
\caption{The flowchart of the model transmission. \textit{Mode 1}: the swarm robots detect the environment, including information on the threats, obstacles, and other robots. \textit{Mode 2}: if one robot detects the threat, this robot records the position of the threat and generates the escape trajectory. \textit{Mode 3}: the resultant force combines the attractive and repulsive forces, where two self-adaptive weights are used to adjust the influence of the attractive and repulsive forces.}
\label{model_trans}
\end{figure}

Fish schools exhibit highly efficient group behaviors through simple individual interactions. In this paper, the escape behavior is modeled as a mode transition process. Each robot has three modes: \textit{Align} mode, \textit{Escape} mode, and \textit{Follow} mode. The behavior of the robot is determined by a current mode. As shown in Fig. \ref{model_trans},  the swarm robots start in the \textit{Align} mode. In the case that the robot has not detected a threat within the sensor range $R_s$ and any mode transitions from neighboring robots, the robot stays in \textit{Align} mode. If the robot detects the threat, its \textit{Align} mode transitions into the \textit{Escape} mode. In the case that the robot detects that any neighboring robots transition into the \textit{Escape} or \textit{Follow} mode, this robot transitions into the \textit{Follow} mode. In the case that the \textit{Follow} robot detects a new threat, this robot transitions into the \textit{Escape} mode, whereas the original \textit{Escape} robot transitions into the \textit{Follow} mode. In the event that the \textit{Escape} robot  achieves the safe distance $d_s$, the escape task is finished.

\begin{figure}[t]
\centering
\includegraphics[width=2.7in]{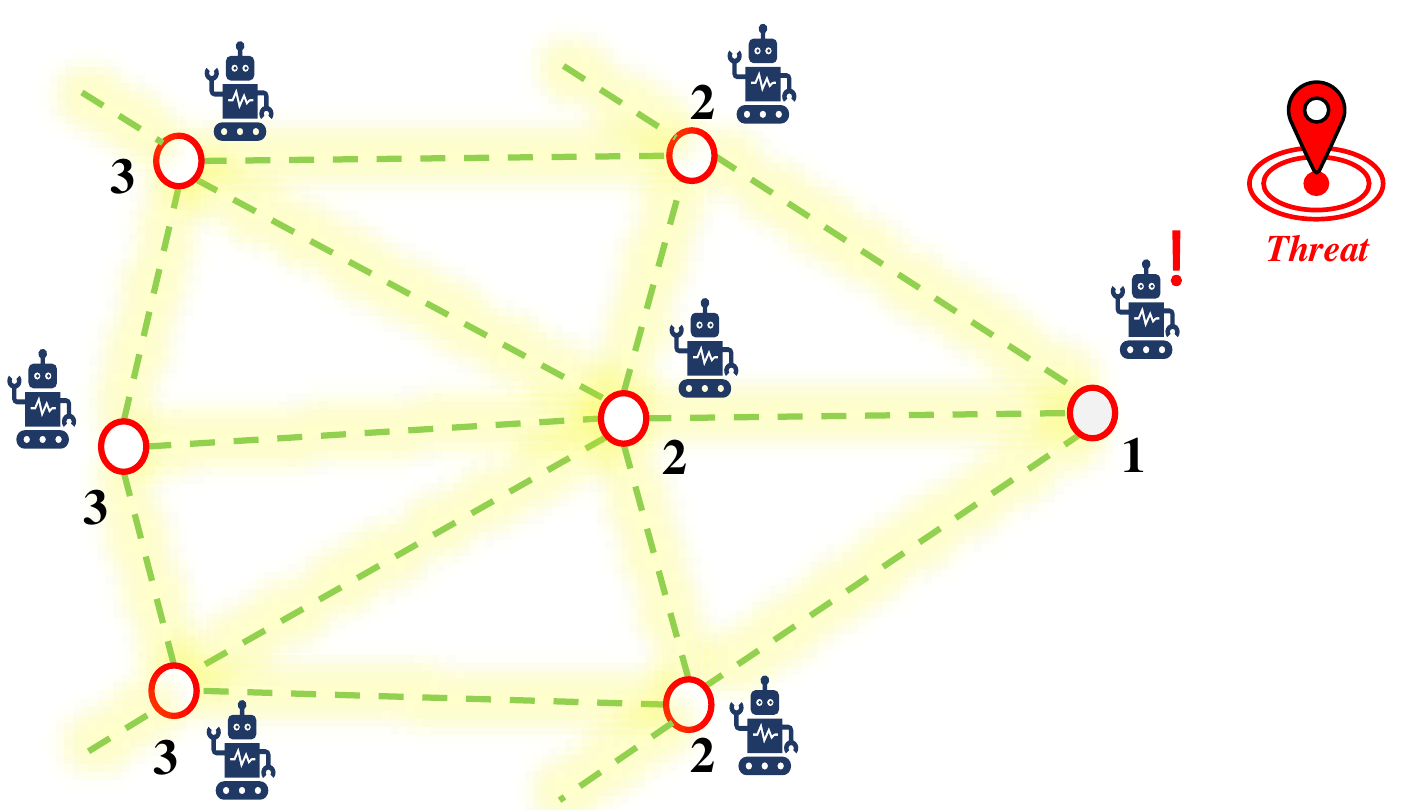}
\caption{The illustration of fish-inspired collective organization. The robots cannot share information with neighbors, only the first robot to detect the threat is aware of the threat position.}
\label{group_map}
\end{figure}

In addition, 
the fish organization was considered an egalitarian organization. Therefore, the traditional fish-inspired approaches assumed that swarm robots are an egalitarian organization \cite{berlinger2021implicit}. However, the most current studies found that the hierarchical organization might exist in some species of fish \cite{ioannou2017swarm}. Therefore,
inspired by Nagy \textit{et al.} \cite{nagy2010hierarchical}, a hierarchical organization is incorporated into the swarm robots. As shown in Fig. \ref{group_map}, once a robot detects the threat, the escape of this robot begins. At this time, a  hierarchical index system is built to organize the swarm robots. The hierarchical index of the \textit{Escape} robot is $1$. The hierarchical index of robots within the detection range of the \textit{Escape} robot is $2$. For robots outside the detection range of \textit{Escape} robot, the hierarchical index is equal to the lowest hierarchical index of its neighbors plus 1. The hierarchical organization can reduce the computation cost when the system considers a large number of robots. Only the \textit{Escape} robot needs to be notified of the position of the threat and dynamically generate the escape trajectory. If the threat is observed by multiple robots, robots that detect threat will transition into the \textit{Escape} mode and multiple escape trajectories will be generated. Other robots will follow the closest \textit{ Escape} robot when observing multiple \textit{Escape} robots within their detection range.


\subsection{Virtual Forces Based on Bio-inspired Neural Network}
\begin{figure}[!t]
\centering
\includegraphics[width=3.4in]{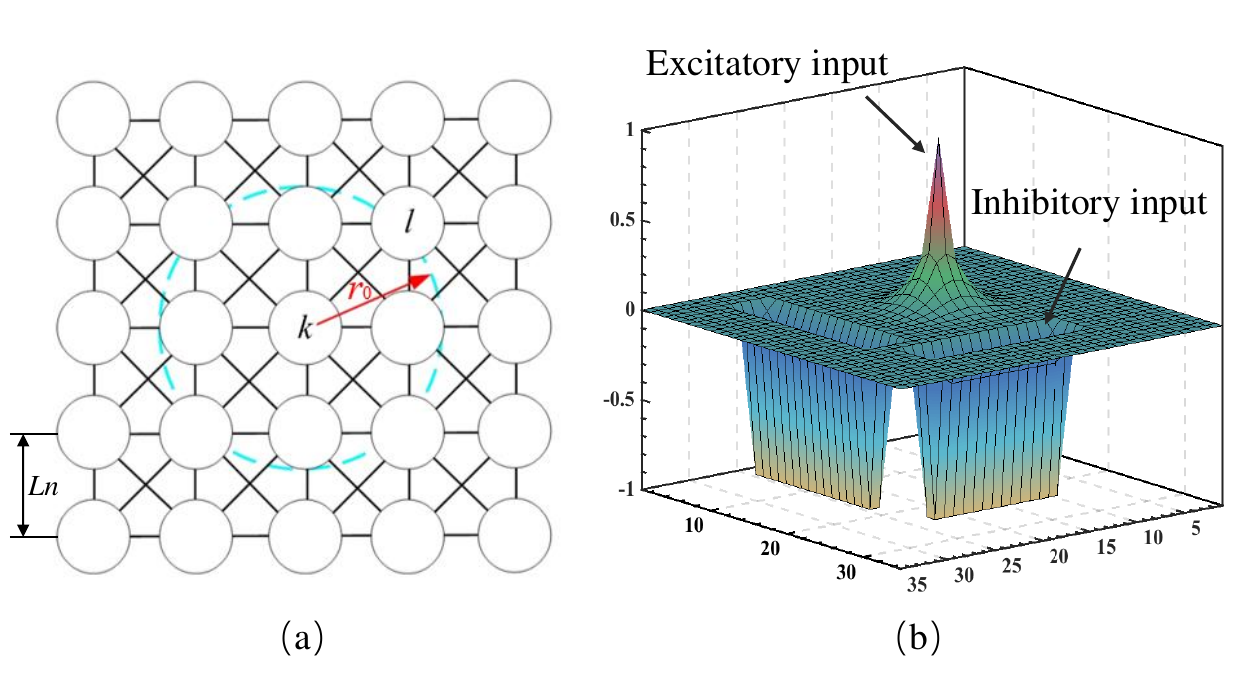}
\caption{ Examples of the bio-inspired neural network. (a) structure of the neural network with only local connections; (b) the dynamic landscape of neural activity.}
\label{fig_BINN}
\end{figure}

During the past decades, there have been many models to mimic fish shoaling \cite{berlinger2021implicit}. Many models consider the virtual forces to be an effect of individual fish \cite{parrish2002self}. In this paper, a BINN is proposed to generate attractive and repulsive forces. 
The structure of the BINN is shown in Fig. \ref{fig_BINN}(a). 
The proposed  neural network architecture is characterized by a discrete topologically organized map. In this configuration, each neuron uniquely represents a specific environmental location. Note that the distance between neurons $L_n$ is assumed to be equal to $\lambda_o$, which can guarantee the neuron one-to-one representing  the obstacle.
Therefore, the receptive field of the  neuron is represented by a circle with a radius of $r_0$ and has lateral connections only to its eight neighboring neurons. 
The fish-inspired organization is used to reduce computational complexity. Only the \textit{Escape} robot needs to use a neural network with the one-to-one environment representing to dynamically generate the escape trajectory. The \textit{Follow} robot can use a smaller neural network to leave the threat.

The shunting model was developed by Grossberg \cite{grossberg1988nonlinear} based on Hodgkin and Huxley's model \cite{hodgkin1952quantitative}. The shunting equation can be written as 
\begin{equation}
\frac{d x_{k}}{d t}=-A x_{k}+\left(B-x_{k}\right) S_{k}^{e}-\left(D+x_{k}\right) S_{k}^{i}
\label{eq:shunnting}
\end{equation}
where $x_{k}$ denotes the neural activity of $k$-th neuron; $S^e_k$ and $S^i_k$ are the excitatory and inhibitory inputs to the neuron, respectively; $A$ is the passive decay rate; $B$, and $D$ are the upper and lower bounds of the neural activity, respectively. Several robotic navigation and control algorithms have been developed depending on the shunting model \cite{Li2021bio}. 
The neural activity for $k$-th neuron is written as
\begin{equation}
\begin{aligned}
\frac{d x_{k}}{d t}=-Ax_k&+(B-x_k)\Bigg([I_k]^++\sum^n_{l=1}w_{kl}[x_l]^+\Bigg)\\
&-(D+x_k)\Bigg([I_k]^-+\sum^n_{l=1}g_{kl}[x_l-\sigma]^-\Bigg)
\end{aligned}
\label{eq:binnE}
\end{equation}
where $x_l$ represents the neural activity of neighboring neurons to $k$-th neuron; $n$ represents the amount of neighboring neurons to $k$-th neuron; $[a]^+$ is defined as $[a]^+=\max\left\{a,0\right\}$; $[a]^-$ is defined as $[a]^-=\max\left\{-a,0\right\}$; and $\sigma$ is the threshold of the inhibitory lateral neural connections. The connection weight, $w_{kl}$ and $g_{kl}$, are defined as 
\begin{equation}
w_{kl}=f(|kl|)=\left\{
\begin{aligned}
&\mu/|kl|,  \quad  0<|kl|\leq r_0\\
&0, \quad \qquad |kl|>r_0
\end{aligned}
\right.
\label{eq:connection}
\end{equation}
and 
\begin{equation}
g_{kl}=\beta w_{kl},
\label{eq:connection_v}
\end{equation}
respectively, where $\beta$ is a positive constant, $\beta \in [0, 1]$; $|kl|$ represents the Euclidean distance between $k$-th neuron to  $l$-th neuron; $\mu$ is a positive constant. 
The excitatory signal $S^e_k$ is designed to the term $[I_k]^++\sum^n_{l=1}w_{kl}[x_l]^+$.  The term $\sum^n_{l=1}w_{kl}[x_l]^+$ calculates the sum of positive neural activities from its neighboring neurons. For each neuron, the computation of its neural activity involves calculations from its neighboring neurons. 
The inhibitory signal $S^i_k$ is designed to the term $[I_k]^-+\sum^n_{l=1}g_{kl}[x_l-\sigma]^-$. The term $\sum^n_{l=1}g_{kl}[x_l-\sigma]^-$ guarantees that negative neural activity can propagate in a small region because the existence of the threshold $\sigma$ affects the inhibitory lateral connections. Only negative neural activity that reaches the threshold can propagate to its neighboring neurons.
Therefore, the excitatory signal $S^e_k$ globally influences the whole state space, while the inhibitory signal $S^i_k$ has only a local effect in a small region, as shown in Fig. \ref{fig_BINN}(b). The external input $I_k$ varies with the generation of force, which can be defined as
\begin{equation}
I_k=\left\{
\begin{aligned}
&I_{att},  \quad \, \text{if generate a attractive force} \\
&I_{rep},  \quad \text{if generate a repulsive force}  \\
\end{aligned}
\right.
\end{equation}
where $I_{att}$ and $I_{rep}$ are external inputs when generating the attractive and repulsive forces, respectively. 
Based on the BINN, the attractive and repulsive virtual forces can be defined as follow.




\subsubsection{Attractive Force}
Based on the modeling of the fish-inspired behavior, the \textit{Follow} robots need to track the \textit{Escape} robot or the lowest hierarchical index \textit{Follow} robot. The lower hierarchical index robot indicates it is closer to the \textit{Escape} robot, which has higher probabilities of the correct moving direction to the \textit{Escape} robot. Thus, the external input $I_{att}$ is defined as
\begin{equation}
I_{att}=\left\{
\begin{aligned}
&E,  \qquad \, \text{if it is a \textit{Escape} or lowest index robot } \\
&-E,  \quad \text{if it is an obstacle}  \\
&0,   \qquad \ \, \text{otherwise}
\end{aligned}
\right.
\end{equation}
where $E$ is a positive constant. If the corresponding position of the neuron is the \textit{Escape} robot or the lowest index robot, the external input becomes a large positive value. If the corresponding position is the obstacle, the external input  becomes a large negative value. The command neuron of the attractive force can be given as
\begin{equation}
\begin{aligned}
P_{att} \Leftarrow x_{P_{att}}=\rm{max}&\left\{ x_l,l=1,2,...,n\right\}&
\label{search_max}
\end{aligned}
\end{equation}
where $P_{att}$ represents the command neuron of the attractive force in the neural network; $x_{P_{att}}$ represents the neural activity of the command neuron of the attractive force. From (\ref{search_max}), the robot keeps searching for maximum neural activity from its neighborhoods. When a robot advances to a new position, the new position becomes its current position.  The attractive force $\overline{ \mathbf{f_{A}(k, l)}}$ can be defined as
\begin{equation}
\overline{\mathbf {f_{A}(k, l)}}=C_A
\frac{{P}_{att}-{P_c}}{\left\|{P_{att}}-P_c\right\|}
\label{attr_force}
\end{equation}
where $C_A$ is a positive constant; $P_c$ is the current position of the robot. Because the neuron of the obstacle has a negative activity value, the robot will not choose the obstacle neuron as the next position. 
Therefore, the attractive force $\overline{ \mathbf{f_{A}(k, l)}}$ is also collision-free to the obstacle.

\subsubsection{Repulsive Force}
Based on the modeling of the fish-inspired behavior, when robots transition into the \textit{Escape} mode, the \textit{Escape} robot needs to leave away from the threat. Meanwhile, the swarm robots are required to maintain the desired distance $R_d$ to each other. Thus, the external input $I_{rep}$ is defined as
\begin{equation}
I_{rep}=\left\{
\begin{aligned}
&E,  \qquad \, \text{if it is a threat or neighbor robot} \\
&-E,  \quad \text{if it is an obstacle}  \\
&0,   \qquad \ \, \text{otherwise}.
\end{aligned}
\right.
\end{equation}
 If the corresponding position of the neuron is a threat or neighbor robot, the external input becomes a large positive value. If the corresponding position is an obstacle, the external input  becomes a large negative value. The command neuron of the repulsive force can be given as
\begin{equation}
\begin{aligned}
P_{rep} \Leftarrow x_{P_{rep}}=\rm{min}&\left\{ x_l,l=1,2,...,n; x_{l} \geq 0\right\}&
\label{search_min}
\end{aligned}
\end{equation}
where $P_{rep}$ represents the command neuron of the robot; $x_{P_{rep}}$ represents the neural activity of the command neuron of the repulsive force.  The repulsive force of the \textit{Follow} robot $\overline{\mathbf{f_{R}(k, l)}}$ can be defined as 
\begin{equation}
\overline{\mathbf{ f_{R}(k, l)}}= \begin{cases}C_R\frac{{P}_{rep}-{P_c}}{\left\|{P_{rep}}-P_c\right\|} & \text { if } 0<D (k, l) \leq R_{d} \\
\overline{\mathbf{0}}, & \text { if } D (k, l)>R_{d}\end{cases}
\label{repu_force}
\end{equation}
where $C_R$ is a positive constant; $D (k, l)$ is the distance between two neighboring robots $k$ and $l$. The repulsive force of the \textit{Follow} robot takes effect only if the distance between the neighboring robots is less than $R_d$. Because the neuron of the obstacle has a negative activity value, the robot will not choose the obstacle neuron as the next position. Therefore, the attractive force $\overline{\mathbf{f_{R}(k, l)}}$ is also collision-free to the obstacle. 
The \textit{Escape} robot has information about the threat position, thus the \textit{Escape} robot only receive the repulsive force from the threat to generate the collision-free escape trajectory.
The repulsive force of the \textit{Escape} robot $\overline{\mathbf{F_{RT}}}$ can be defined as 
\begin{equation}
\overline{\mathbf{ F_{RT}}}= \begin{cases}C_R\frac{{P}_{rep}-{P_c}}{\left\|{P_{rep}}-P_c\right\|} & \text { if } 0<D (i,Th) \leq d_{s} \\
\overline{\mathbf{0}}, & \text { if } D (i,Th)>d_{s}\end{cases}
\label{repul_force}
\end{equation}
where $D (i,Th)$ is the distance between the $i$-th \textit{Escape} robot and threat. The repulsive force of the \textit{Escape} robot takes effect only if the distance is smaller than the safe distance $d_s$.
The stability and convergence of the proposed model can be rigorously proved using the Lyapunov stability theory.
Equation (\ref{eq:binnE}) can be written into Grossberg's general form\cite{grossberg1988nonlinear},
\begin{equation}
\frac{d x_{k}}{d t}=a_{k}\left(x_{k}\right)\left(b_{k}\left(x_{k}\right)-\sum_{l=1}^{n} c_{k l} d_{l}\left(x_{l}\right)\right)
\label{eq:general}
\end{equation}
based on the substitutions as follows:
\begin{equation}
a_{k}\left(x_{k}\right)=\left\{
\begin{aligned}
&B-x_{k} \quad \text{if}  \ x_{l} \geq 0\\
&D+x_{k} \quad \text{if}  \ x_{l} < 0
\end{aligned}
\right.
\end{equation}
\begin{equation}
\begin{aligned}
c_{k l}=&-w_{k l} \\
\end{aligned}
\end{equation}
\begin{equation}
d_{l}\left(x_{l}\right) =\left\{
\begin{aligned}
&x_{l} \qquad \qquad   \text{if}  \ x_{l} \geq 0\\
&\beta(x_{l}-\sigma)\quad  \text{if}  \ x_{l} < \sigma\\
&0 \qquad \qquad  \  \enspace \text{otherwise}
\end{aligned}
\right.
\end{equation}
\begin{equation}
\begin{aligned}
b_{k}\left(x_{k}\right)=& \frac{\left(B\left[I_{k}\right]^{+}-D\left[I_{k}\right]^{-}-(A+\left[I_{k}\right]^{+}+\left[I_{k}\right]^{-})x_{k}\right)}{a_{k}\left(x_{k}\right)}.
\end{aligned}
\end{equation}
Since $w_{k l}=w_{l k}$,  it follows that $c_{k l}=c_{l k}$, indicating symmetry. Since $x_{k}$ is bounded in the interval $[-D, B]$, it follows that $a_{k}\left(x_{k}\right) \geq 0$, indicating positivity. Since  $d_{l}^{\prime}\left(x_{l}\right)=1$ at $x_{l}>0$; $d_{l}^{\prime}\left(x_{l}\right)=\beta \geq 0$ at $x_{l}<\sigma$; and $d_{l}^{\prime}\left(x_{l}\right)=0$, otherwise, it follows that  $d_{l}^{\prime}\left(x_{l}\right)\geq 0$, indicating monotonicity. Therefore, (\ref{eq:binnE}) satisfies all the three stability conditions (symmetry, positivity, and monotonicity) required by Grossberg's general form  \cite{grossberg1988nonlinear}. The candidate of the Lyapunov function for (\ref{eq:general}) can be chosen as \cite{cohen1983absolute}
\begin{equation}
\begin{aligned}
V=&-\sum_{k=1}^{n} \int^{x_{k}}_0 b_{k}\left(\xi_{k}\right) d_{k}^{\prime}\left(\xi_{k}\right) d\xi_{k}\\
&+\frac{1}{2} \sum_{l, j=1}^{n} c_{l j} d_{l}\left(x_{l}\right) d_{j}\left(x_{j}\right).
\end{aligned}
\end{equation}
For the generated force, the time derivative of $V$ can be given by
\begin{equation}
\frac{d V}{d t}=-\sum_{k=1}^{n} a_{k} d_{k}^{\prime}\left(b_{k}-\sum_{l=1}^{n} c_{k l} d_{l}\right)^{2}.
\end{equation}
During the generation process of the virtual force, $d V/{d t} \leq 0$ because of $a_{k}\geq 0$ and $d_{k}^{\prime} \geq 0$. The more detail and rigorous proof can be found in \cite{cohen1983absolute}. As a result, the proposed model is stable. The dynamics of the neural network is guaranteed to converge to an equilibrium state of the system.

\subsection{Self-adaptive Collective Mechanism}

The proposed BINN-based virtual forces enable robots to leave the threat with collision avoidance. During the escape process, the swarm robots are required to be self-adaptive to adapt to dynamic environments, which means the swarm robots should be able to dynamically adjust their movement parameters based on the environment.  The resultant force of each robot can be given as
\begin{equation}
\overline{\mathbf{F_{{RS}}}}=\alpha_{A}\sum_{l\in N(I), h_{l}<h_{k}}\overline{\mathbf{f_{A}(k, l)}}+\alpha_{R} \sum_{l \in N(I)} \overline{\mathbf{f_{R}(k, l)}}
\end{equation}
where $\alpha_{A}$ and $\alpha_{R}$, $ 0 \leq \alpha_{A},\alpha_{R} \leq 1$ and $ \alpha_{A} + \alpha_{R} = 1$, are self-adaptive weights of the attractive and repulsive forces, respectively. The self-adaptive motion is to dynamically adjust the proper ratio of $\alpha_{A}$ / $\alpha_{R}$ to adapt the environmental changes.
In the proposed method, the dynamic neural activity is incorporated into the adjustment of the $\alpha_{A}$ / $\alpha_{R}$ ratio, as shown in lines 7 to 10 in Algorithm \ref{alg:Self-adaptive}. 
The stride lengths of adjustment $\Delta$ can be defined as
\begin{equation}
 \Delta =\left\{ 
\begin{aligned}
 &+U, \qquad \, \text{if $ \mathrm{Avr(i)} + \sum^n_{l=1}[x_l]^->R_d$}   \\
& -U,  \qquad \, \text{if $\mathrm{Avr(i)} + \sum^n_{l=1}[x_l]^- \leq R_d$}   \\
\end{aligned}
\right.
	\label{ratioadjust}
\end{equation}
where $U$ is a small constant.
The function $\mathrm{Avr(i)}$ denotes the average neighboring distance of the robot $i$. Function $[a]^-=\max\left\{-a,0\right\}$ denotes the sum of neighboring negative neural activity. If no obstacles are in proximity to the robots, the term representing neural activity effect, $\sum^n_{l=1}[x_l]^-$, should equate to $0$. Therefore, robots adjust the ratio depending on whether their average neighboring distance is less than or exceeds $Rd$. If there are obstacles in proximity to the robots, $\sum^n_{l=1}[x_l]^-$ is a large positive value. Thus, the attraction effect continuously increases, which ensures that robots remain connected with each other to bypass obstacles. A detailed discussion of neurodynamics-based self-adaptive motion can be found in Section \uppercase\expandafter{\romannumeral6}-B.
\begin{algorithm}[t]
	\LinesNumbered
	\caption{Neurodynamics-based Self-adaptive mechanism}
	\label{alg:Self-adaptive}
	
	\KwIn{positions of robot $\mathbf{p}_{e}= (x_e,y_e)$, desired distance $R_d$ and dynamic neural activity $N$}
	\KwOut{ratios of $\alpha_{A}$ / $\alpha_{R}$}
 
	Initialize stride lengths of adjustment $\Delta$\\
	$\alpha_{A}$ $\rightarrow$ 0.5\\
	$\alpha_{R}$ $\rightarrow$ 0.5\\
 
	\While{escape from the threat}
	{
		\If{$0\leq \alpha_{A} \leq 1 $ and $0 \leq \alpha_{R}\leq 1$}
		{
			    get neighbouring neural activity $x_l$ from $N$\\
			\If{$\mathrm{Avr(i)}+ \sum^n_{l=1}[x_l]^->R_d$}
			{
				increase ratio $\alpha_{A}$ / $\alpha_{R}$ by $\Delta$\\
			}	
			\Else
			{
				decrease ratio $\alpha_{A}$ / $\alpha_{R}$ by $\Delta$\\
			}
		}
	}
 	
\end{algorithm}

The virtual forces are used to determine the collision-free position of the next movement. However, when to move to the next position is determined by the velocity of the robot. The magnitude of $\overline{\mathbf{F_{{RS}}}}$ is used to determine the velocity of the robot $v_{k}$. The possible magnitude of $\overline{\mathbf{F_{{RS}}}}$ ranges from $0$ to $+\infty$, whereas the $v_{k}$ is a limited range from $0$ to the maximum velocity $V_{max}$. Thus, the magnitude of $\overline{\mathbf{F_{{RS}}}}$ is required to map into a finite velocity. The velocity of the robot $v_k$ is given by
\begin{equation}
v_{k}=\arctan \left(\left|\overline{\mathbf{F_{{RS}}}}\right|\right) \times(2 / \pi) \times V_{\max }
\end{equation}
where $\arctan()$ is the trigonometric function. The above nonlinear mapping has been used for the collective motion \cite{zhao2018self}.  The velocity $v_{k}$ increases with the increase of the magnitude of $\overline{\mathbf{F_{{RS}}}}$ until the force $\overline{\mathbf{F_{{RS}}}}$ reaches a larger magnitude, which is less sensitive to the mapping function. For the \textit{Escape} robot, the same method is employed to map $\overline{\mathbf{F_{{RT}}}}$ onto velocity $v_{k}$.

\section{Simulation Results}
\label{sec:simulation}
\begin{figure*}[!t]
\centering
\includegraphics[width=7in]{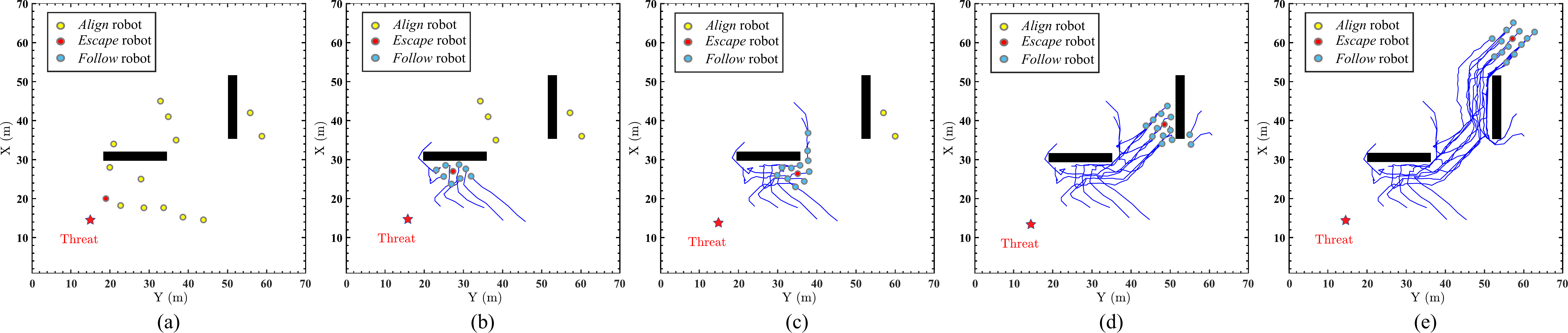}
\caption{Robots escape in a static environment. (a) the initial position of robots; (b) robots escape at time $8s$; (c) robots escape at 16s; (d) robots escape at $43s$; (e) robots escape at $52s$.}
\label{fig_staticObs}
\end{figure*}

\begin{figure*}[t]
\centering
\includegraphics[width=7in]{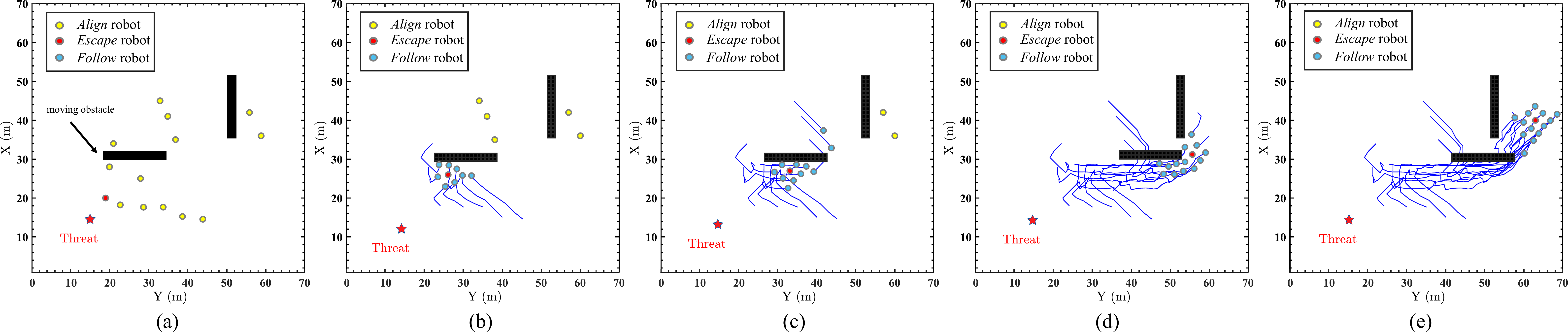}
\caption{Robots escape in a dynamic environment. (a) the initial position of robots and moving obstacles; (b) robots escape at $7s$; (c) robots escape at $14s$; (d) robots escape at $36s$; (e) robots escape at $45s$.}
\label{fig_dynamic}
\end{figure*}

\begin{figure*}[t]
\centering
\includegraphics[width=7in]{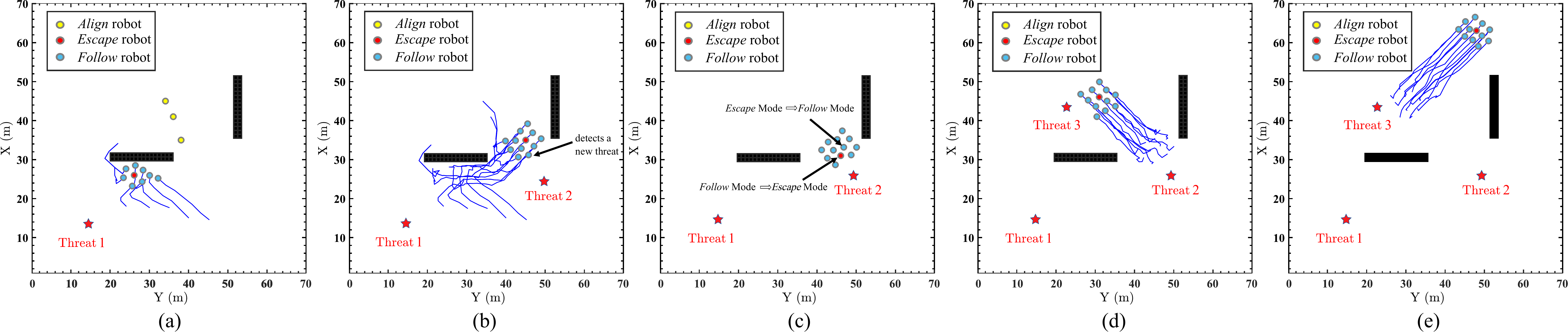}
\caption{Robots escape in a sudden change environment. (a) robots detect the threat and escape at $6s$;  (b) a new threat is detected at $25s$; (c) two robots transition their modes at time $26s$; (d) a new threat is detected at $41s$; (d)robots escape at $61s$.}
\label{fig_sudden}
\end{figure*}
 In this section, the proposed approach is tested to a series of evaluations across various scenarios \cite{Li_2019}. 
 In addition, simulation results are compared with other recent approaches to illustrate the performance of the proposed approach. All simulation studies are tested in MATLAB R2021a. The swarm robots are randomly distributed in the environment. The parameters of all simulations are set as follows: $A= 15$, $B= 1$, $D=1$, $\mu= 1$, $E= 70$, $\sigma=-0.5$, $L_n=1$, $r_0=\sqrt{2}$, $R_d= 3$, $R_s=8$ and $V_{max}=1.4$. The environment is represented by a neural network, which has $70 \times 70$ neurons.

\subsection{Escape with Static Obstacles}

The first simulation aims to test the proposed approach with the static obstacle. As shown in Fig. \ref{fig_staticObs}, there are 13 robots randomly deployed in the environment, where the position of the threat is (15,15). In the beginning, one robot detects the threat and transmission to the \textit{Escape} mode (red color), as shown in Fig. \ref{fig_staticObs}(a). As the \textit{Escape} robot moves, the individual robots detect the \textit{Escape} robot and transmission from the \textit{Align} mode (yellow color) to  the \textit{Follow} mode (blue color), as shown in Figs. \ref{fig_staticObs}(b)–(e). In this process, the individual robots can be dynamically added to the swarm without the need for explicit reorganization, as shown in Figs. \ref{fig_staticObs}(c) and \ref{fig_staticObs}(d). 
 It is important to note that the swarm topology is not the same as in the original, but it still keeps the desired distance $R_d$ between neighbors. 

\subsection{Escape with Moving Obstacles}

In the next simulation, the moving obstacles are considered in the environment. Fig.\ref{fig_dynamic}(a) shows the moving obstacle, which continuously moves to the left.  The moving obstacle might partition the swarm robots, which requires robots to maintain a connection with each other to avoid member loss. Figs. \ref{fig_dynamic}(b)-(e) show the swarm robots are able to bypass the moving obstacle without member loss. A different escape trajectory is generated compared with the last simulation because the moving obstacle blocks the escape direction of the swarm robot. Note that the generation of the new escape trajectory is only based on the dynamic change of neural activity without any learning or decision-making process.

\subsection{Escape with New Threats}

In this simulation, some new threats might suddenly appear in the environment. The mode transition is dynamic because every robot that detects the new threat can transition into the \textit{Escape} mode. As shown in Figs. \ref{fig_sudden}(a) and \ref{fig_sudden}(b), robots start to escape the threat until a new Threat 2 is detected at time 25s. The robot with \textit{Escape} mode transitions into the \textit{Follow} mode, whereas the robot with \textit{Follow} mode transitions into the \textit{Escape} mode, as shown in Fig. \ref{fig_sudden}(c). The new collision-free escape trajectories are generated. Note that the influence of Threats 1 and 2 exists continuously in the neural network. Thus, the swarm robots will not return to Threats 1 or 2 when another threat appears, as shown in Fig. \ref{fig_sudden}(e).

\subsection{Comparison Studies}
In comparison studies, the proposed approach is compared with three recent methods to evaluate the performance of the proposed approach. 
To evaluate the performance of the proposed approach, a total of 30 test cases are conducted in each evaluation. In these tests, the positions of robots, threats, and obstacles are randomly distributed in the environment.

\subsubsection{Environmental Adaptation}
\begin{figure}[t]
\centering
\includegraphics[width=3in]{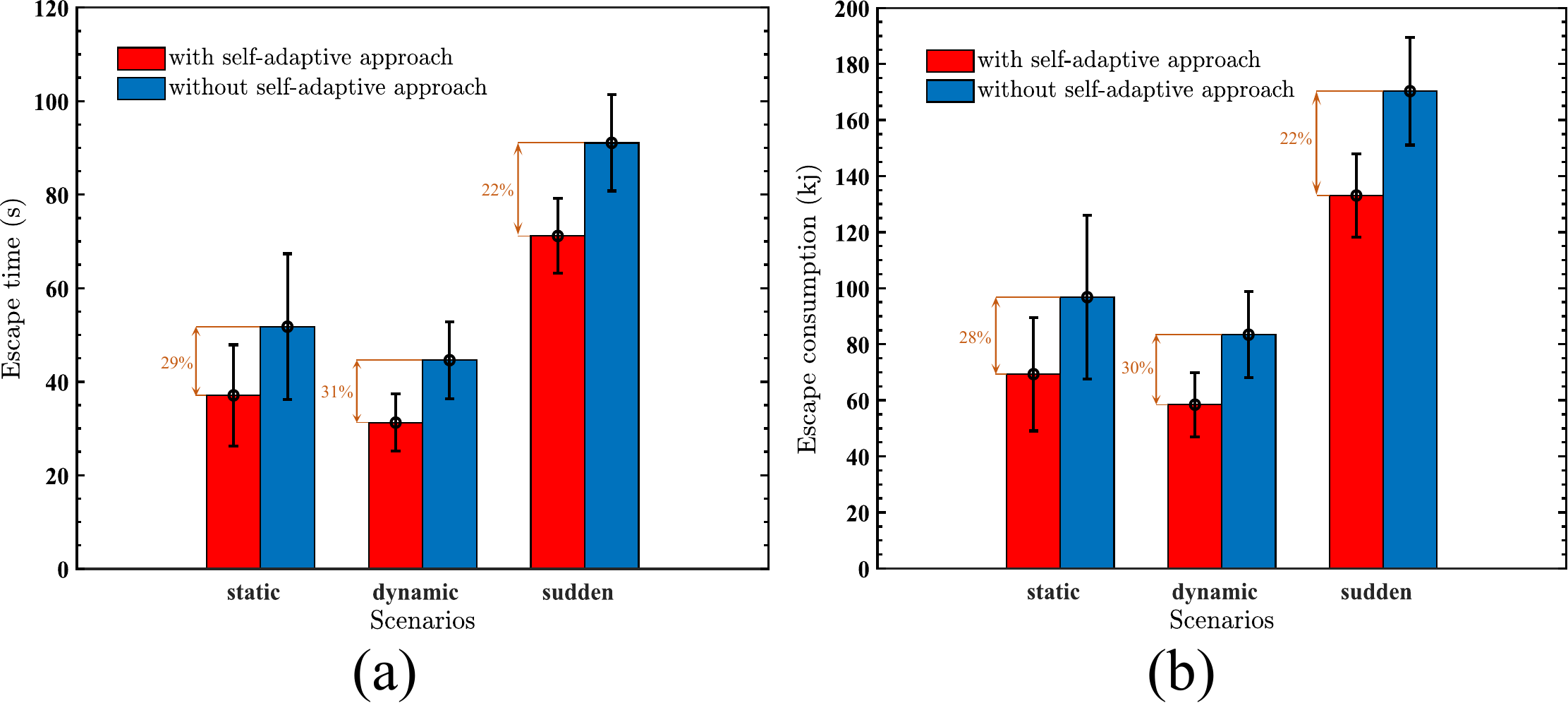}
\caption{Comparisons of escape time and energy consumption in different scenarios. (a) escape time; (b) energy consumption.}
\label{fig:compare_studies}
\end{figure}

The self-adaptive process has more benefits in environmental adaptation.
The proposed approach is compared to an ablative version that lacks the self-adaptive mechanism. The self-adaptive ratio, $\alpha_{A} / \alpha_{R}$, is set to a constant without changes during the escape process. The results, shown in Fig. \ref{fig:compare_studies}, demonstrate that the proposed approach is more effective in all scenarios, particularly in the moving obstacle scenario, where it saves over $31\%$ of the escape time and $30\%$ of the escape consumption. This highlights the importance of the self-adaptive mechanism in providing effective escape solutions.
In addition, the state-of-the-art methods are compared with the proposed approach in the moving obstacles scenario, as shown in Table \ref{table_compare}.

In comparison to  Berlinger \textit{et al.}'s method, Berlinger \textit{et al.}'s method is limited in its ability to provide efficient escape solutions because it only considers a collision-free environment. As shown in Table \ref{table_compare}, the success rate of this approach is low because many robots collide with obstacles. Additionally, the escape trajectory generated by Berlinger \textit{et al.}'s  method is inefficient as it requires the robots to maintain a constant angle to the threat, resulting in wasted energy and time. This method may be able to mimic fish escape behavior, but it has limitations in its practical application.


In comparison to  Novák \textit{et al.}'s method, Novák \textit{et al.}'s method is limited in its ability to provide efficient escape solutions because they treated both the threat and dynamic obstacles as a same influence. In contrast, the proposed BINN considers the threat and multiple obstacles, including both static and moving obstacles, by using two external inputs, $S^e_k$, and $S^i_k$. This allows the proposed apporach to provide more accurate and effective escape solutions. As shown in Table \ref{table_compare}, the success rate of the proposed approach is significantly higher than that of Novák \textit{et al.}'s method. This is because the escape force generated by Novák \textit{et al.}'s method may cancel each other out in multiple obstacles environment, leading to the disorientation and loss of some members within the swarm during the evasion process.


\subsubsection{Self-Adaptive Performance}
The proposed self-adaptive escape of swarm robots has been compared with Zhao \textit{et al.}'s method, which is based on the average distance of neighboring robots, in a scenario involving moving obstacles. The results, shown in Table \ref{table_compare}, indicate that the proposed approach is more efficient in terms of escape consumption and time. Note that there is a significant difference in escape time. The proposed approach results in an escape time of $42.7s$, while Zhao \textit{et al.}'s method results in an escape time of $62.5s$. This difference is due to the fact that Zhao \textit{et al.}'s method involves a decrease in moving speed when robots bypass obstacles, as the support force from the obstacle cancels out the virtual force perpendicular to the obstacle. In contrast, the proposed approach relies only on dynamic neural activity for obstacle avoidance, allowing for a constant moving speed during the escape process.

\begin{table}[htbp]   
\centering
\caption{The comparison of escape performance to the state-of-the-art methods in the moving obstacles scenario} 
\scalebox{0.93}{
\begin{tabular}{cccc}    
\toprule
  Method & Escape Consumption & Escape Time & Success Rate\\    
\midrule   
Berlinger \textit{et al.} \cite{berlinger2021self}  & $216.71kJ$ & $87.2s$ & $31\%$\\  
Novák \textit{et al.} \cite{novak2021fast}  & $132.12kJ$ & $50.1s$ & $46\%$\\  Zhao \textit{et al.} \cite{zhao2018self}  & $125.62kJ$ & $62.5s$ & $77\%$\\  
Proposed approach & $116.45kJ$  &  $42.7s$& $100\%$\\  
\bottomrule   
\label{table_compare}
\end{tabular} 
}
\end{table}


\section{Experiments}
\label{sec:experiments}

\begin{figure}[t]
\centering
\includegraphics[width=3in]{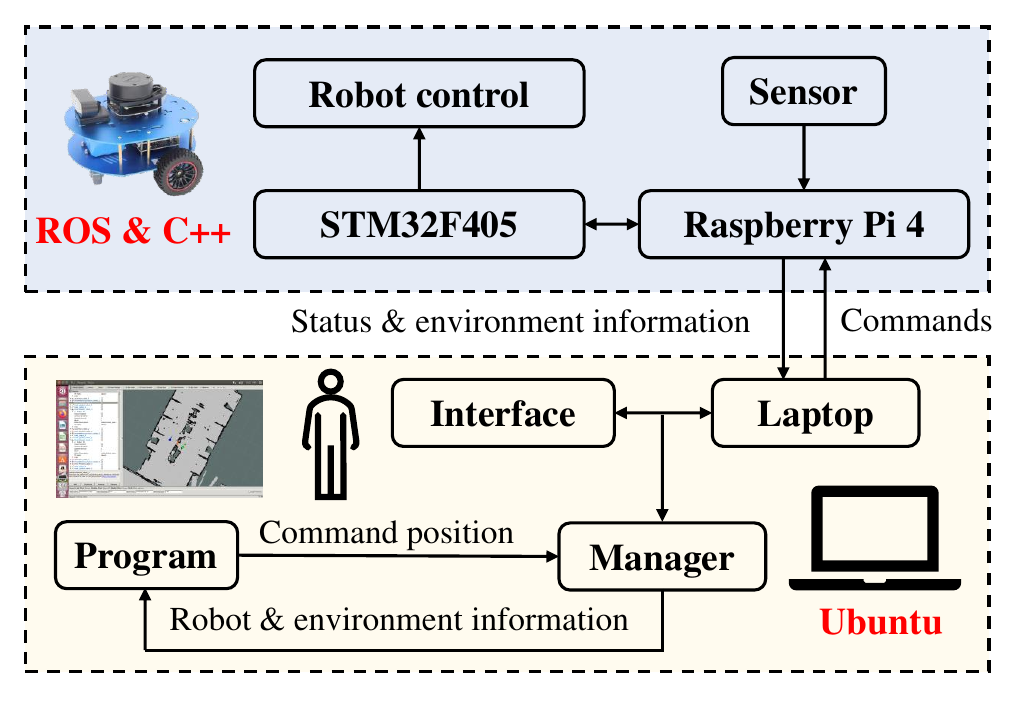}
\caption{The architecture of the real robot test. Blue dotted box: ROS-based mobile robot platform. Yellow dotted box: program implemented in the Ubuntu system.}
\label{fig_logic_real}
\end{figure}

\begin{figure*}[t]
\centering
\includegraphics[width=5.5in]{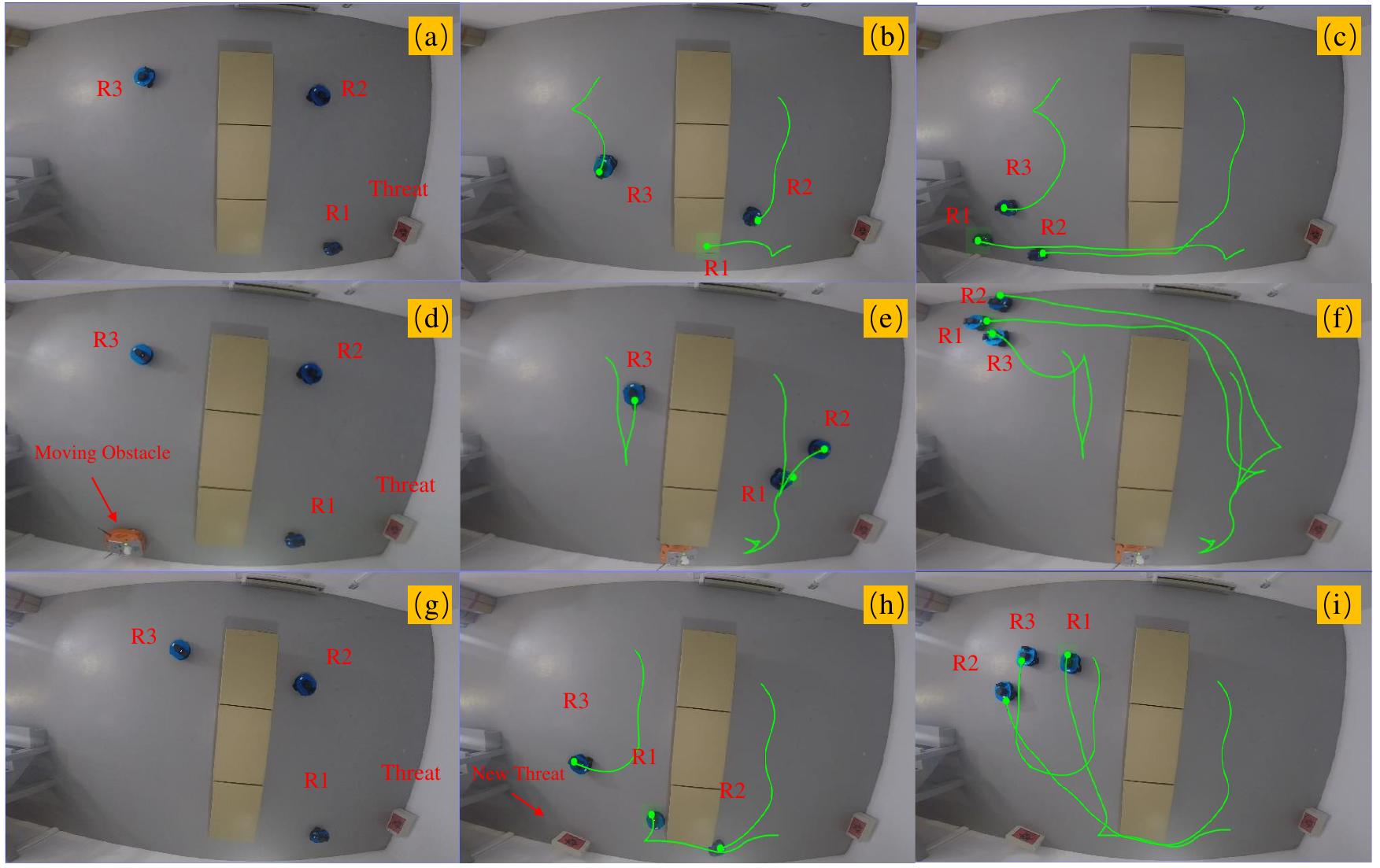}
\caption{Mobile robots experiments in different scenarios. (a) the initial positions of the static obstacle experiment; (b) static environment at time $5.5s$; (c)  static environment at time $15.5s$; (d) the initial positions of the moving obstacle experiment; (e)  dynamic environment at time $9s$; (f) dynamic environment at time $21s$; (g) the initial positions of the new threat experiment; (h) new threat environment at time $9.5s$; (i) new threat environment at time $20.5s$.}
\label{fig_realrobot}
\end{figure*}

The proposed approach was tested using multiple mobile robots in a real-world environment. The architecture of the test platform is shown in Fig. \ref{fig_logic_real}. Three ROS (Robot Operating System)-based robots were built for experimental purposes.
Each robot carried a 1080p camera implemented in Ubuntu using the OpenCV library and one RPLIDAR A1 laser scanner, which has $12m$ and $360\degree$ omnidirectional range scanning.  In addition, the mobile robot contains Raspberry Pi 4 Model B and STM32F405 computing boards for robot control, environment detection, and localization. Mobile robots can send their status and environmental information to the computer using a WIFI network. The proposed approach is implemented on the Ubuntu system, and the Rviz program was used to interface with mobile robots.
The experiment is within an area of $5m$ $\times$ $3.5m$ where robot movements were recorded through a 1080p camera, connected by the WIFI network to mobile devices, and localized on the top of the area. Three polyhedral obstacles are considered, and the threat is chosen as a box with biohazard labeling. Robot dynamics is described by means of differential drive models. 
The proposed velocity $v_k$ can be regarded as the cooperative path planning for robots. When robots detect the threat using the sensor, the command position $x_i = (x_R, y_R,\theta_R)$ is sent to mobile robots based on the changing environment and current position of the robots. 

The experimental results are collected in Fig. \ref{fig_realrobot}. The green lines show the escape trajectories of three mobile robots. Firstly, it is important to note that mobile robots are able to finish escape tasks in all scenarios. Moreover, mobile robots are able to adjust the escape trajectory with the changing environments, as shown in Figs. \ref{fig_realrobot}(d)-(f) and \ref{fig_realrobot}(g)-(h). 
In particular,  when robot R1 detects the threat, mobile robots R1 and R2 move under the obstacle to escape from the threat, as shown in Figs. \ref{fig_realrobot}(b) and \ref{fig_realrobot}(c). In the next experiment, a moving obstacle (an orange color robot) moves to the position under the obstacle and blocks the escape directions of mobile robots, as shown in Figs. \ref{fig_realrobot}(d) and \ref{fig_realrobot}(e). Since mobile robots are not able to pass the obstacle through the trajectory in Figs. \ref{fig_realrobot}(c), new trajectories are generated as shown in Figs. \ref{fig_realrobot}(e) and \ref{fig_realrobot}(f). The dynamic response of mobile robots to changing environments can be better understood by comparing the new threat in Figs. \ref{fig_realrobot}(g)-(i). Since a new threat suddenly adds to the environment, robot R3 transitions into the \textit{Escape} mode, and robot R1 transitions into the \textit{Follow} mode. The new trajectory of robot R3 is to escape the threat. In real-robot experiments, each robot moves in a safe state trajectory due to the self-adaptive mechanism and the threshold of inhibitory connection $\sigma$. In summary, the results of the experiment show that the proposed approach is capable of guiding mobile robots to keep safe navigation and successful completion of escape tasks in changing environments. 

\section{Discussions}
\label{sec:discussion}
  In this section, the characteristic of the neurodynamics model is analyzed. In addition, the reason that the neurodynamics model can improve self-adaptive motion is also discussed.

 \subsection{Collective Escape via Neurodynamics}

   
In order to investigate the escape performance with respect of the neurodynamics model, three important parameters are analyzed. Parameter $A$ is the passive decay rate, which solely determines the transient response of the external input signal.  
To analyze the influence of parameter $A$, several experiments are tested with the same parameter settings, except that $A$ has different values. The energy consumption, escape time, and success rate of swarm robots at $A = 1,5,15,20,$ and $40$ are listed in Table \ref{table_para_A}. The results in Table \ref{table_para_A} show the standard deviation of the escape time ($0.95s$) and energy consumption ($1.45kJ$) at different parameters $A$ are both very small in the case that  all robots can escape the threat. 

\begin{table}[htbp]   
\centering
\caption{Escape performances with different $A$ values}
\label{table_para_A}   
\begin{tabular}{cccc}    
\toprule
Parameter $A$ & Energy Consumption& Escape Time & Success Rate\\    
\midrule   
$A$ =1 & $118.51kJ$ & $42.4s$ & $100\%$\\   
$A$ =5 & $121.55kJ$ & $44.6s$ & $100\%$\\   
$A$ =10 & $120.20kJ$  & $43.8s$ & $100\%$\\ 
$A$ =15 & $117.00kJ$  & $41.7s$ & $100\%$\\ 
$A$ =20 & $119.26kJ$ & $43.1s$ & $100\%$\\   
$A$ =40 & $118.31kJ$  & $43.7s$ & $100\%$\\  
\bottomrule   
\end{tabular}  
\end{table}

Parameter $\mu$ is the weight of the local connection. While the local connection with each neuron is a small region, the propagation of the positive neural activity is able to arrive at the whole neural network. To analyze the influence of parameter $\mu$, several experiments are tested with the same parameter settings, except that $\mu$ has a different value. The escape time, energy consumption, and success rate of swarm robots at $\mu = 0.1,0.5,1$ and $5$ are listed in Table \ref{table_para_mu}. The results in Table \ref{table_para_mu} show that the standard deviation of escape time ($1.03s$) and the energy consumption ($2.04kJ$) is very small at different parameters when all robots can succeed in escaping the threat. 
In addition, when parameter $\mu > 1$, it is very easy to lose some robots  because the propagated activity is amplified.  Thus, the value of $\mu$ is normally selected in the interval $\mu \in (0,1]$.

\begin{table}[htbp]   
\centering
\caption{Escape performances with different $\mu$ values}
\label{table_para_mu}   
\begin{tabular}{cccc}    
\toprule
Parameter $\mu$ & Escape Consumption& Escape Time & Success Rate\\    
\midrule   
$\mu$ =0.1 & $118.94kJ$ & $45.3$ & $100\%$\\ 
$\mu$ =0.5 & $117.18kJ$ & $44.8$ & $100\%$\\  
$\mu$ =1 & $114.02kJ$  & $42.9$ & $100\%$\\ 
$\mu$ =5 & $120.59kJ$  & $45.1$ & $62\%$\\ 

\bottomrule   
\end{tabular}  
\end{table}

\begin{figure}[t]
\centering
\includegraphics[width=2in]{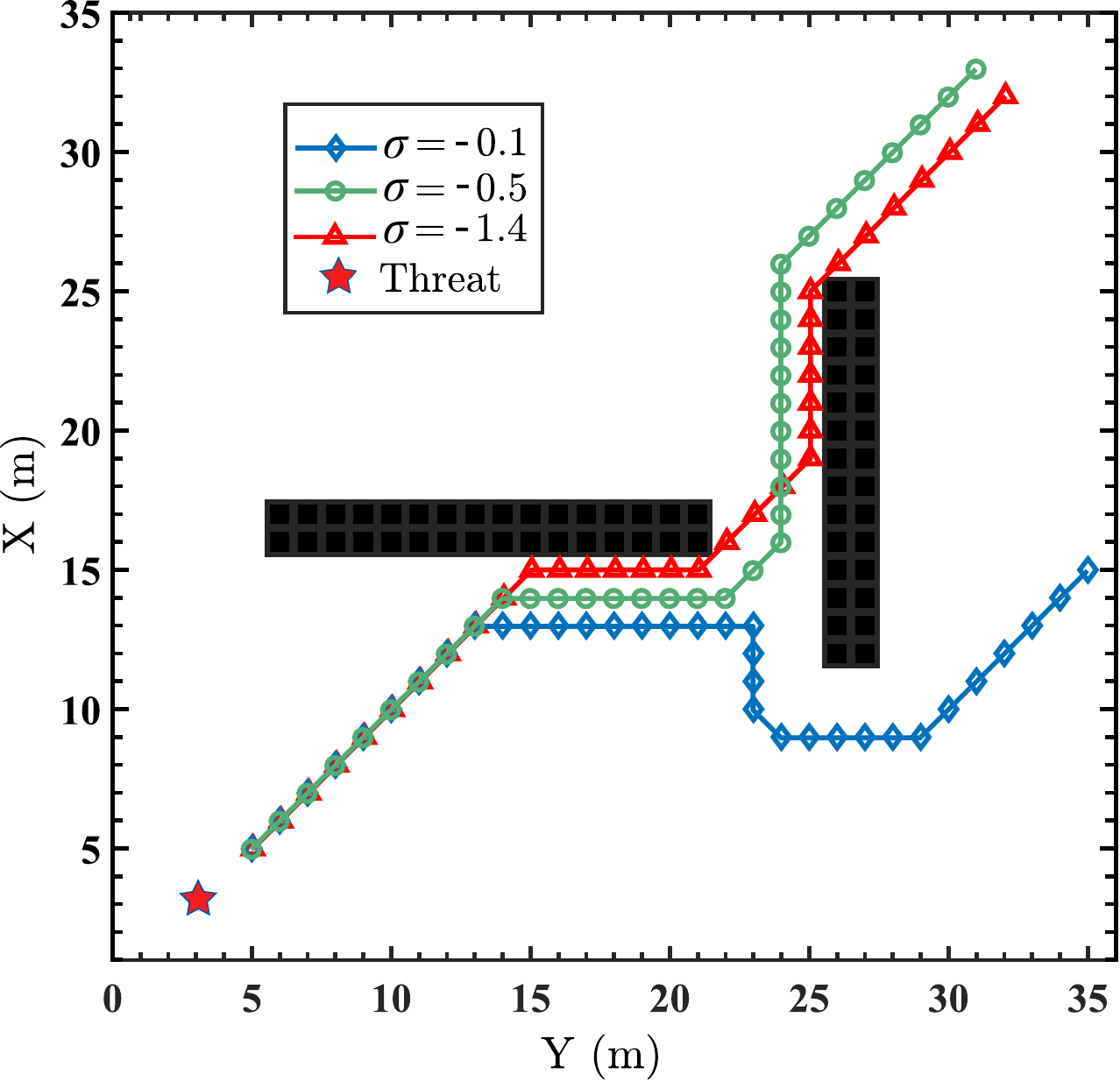}
\caption{Three typical examples of one robot by choosing different parameters $\sigma$ values. }
\label{fig_sigma}
\end{figure}

Parameter $\sigma$ is the threshold of the inhibitory connection, which solely denotes the clearance from obstacles. The obstacle has only local effects in a small region to avoid possible collisions, which can be adjusted by selecting the relative lateral connection strength $\beta$ and the threshold $\sigma$. Fig. \ref{fig_sigma} shows three typical examples of one robot by choosing different parameters $\sigma$. The red and triangular line shows the motion of the robot when choosing $ \sigma= -1.4$. There is no clearance from the obstacles. It is obvious to see that the robot clips the corners of obstacles and runs down the edges of obstacles.  The blue and rhomboid line shows a strong obstacle clearance by choosing $\sigma = -0.1$. It shows a very strong clearance from obstacles, which pushes the robot very far from the obstacle. When considering a larger number of robots or the gap between obstacles is not able to accommodate robots, choosing a small $\sigma$ value is a sub-optimal option. The green and circle line shows the robot motion with a moderate obstacle clearance by choosing $\sigma = -0.5$.  It shows a “comfortable” trajectory, which does not clip the corners of obstacles and runs down the edges of obstacles.

\begin{figure}[htbp]
\centering
\includegraphics[width=3in]{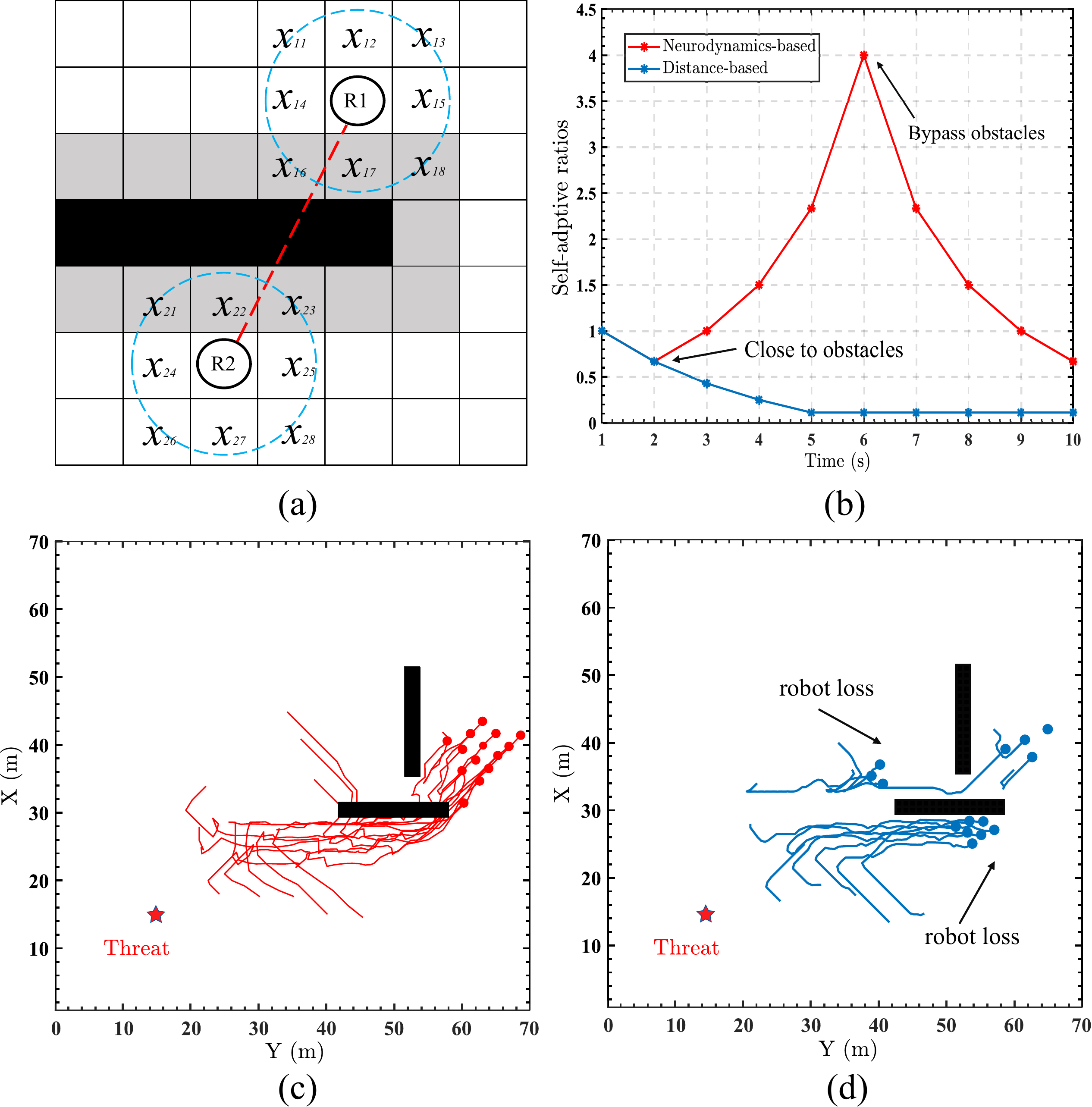}

\caption{The comparison of different self-adaptive motion approaches. (a) the illustration of the neurodynamics-based approach; (b) the comparison of self-adaptive ratio changes. (c) final escape trajectories based on the neurodynamics-based approach; (d) final escape trajectories based on the distance-based approach.}
\label{fig_discon}
\end{figure}

\subsection{Self-adaptive Motion via Neurodynamics}
In traditional approaches to self-adaptive adjustment, the self-adaptive ratio is typically based on the average straight-line distance between neighboring robots \cite{zhao2018self}. The ratio is increased or decreased when the average distance is greater or less than the desired distance. However, the traditional method might not be suitable for the moving-obstacle environments. As illustrated in Fig.\ref{fig_discon}(a), the average distance between two robots (shown in red and dotted line) is less than the desired distance, indicating that the self-adaptive ratio should be decreased. However, in the presence of obstacles that might partition the robots, a decrease in the attraction effect leads to disconnection from other robots. In the proposed approach, the ratio adjustment incorporates neural activity. As shown in Fig.\ref{fig_discon}(a), the obstacle and its clearance have a very large negative neural activity. Neural activity $x_{16},x_{17},x_{18}$ are negative results by (\ref{eq:binnE}), whereas others are positive in neighboring neurons of robot R1. Thus, the term ${\rm Avr(i)} + \sum^n_{l=1}[x_l]^-$  in (\ref{ratioadjust}) should be greater than the desired distance $R_d$. Therefore, the ratio should increase, which increases the effect of the attractive force acting and avoiding the loss of robots. Same with robot R2, neural activity $x_{21},x_{22},x_{23}$ are negative resulted by (\ref{eq:binnE}), whereas others are positive in the neighboring neurons of robot R2. As shown in Fig.\ref{fig_discon}(b), the comparison of ratio changes shows that the proposed approach increases the attractive effect, while the distance-based method decreases the attractive effect when the robots are close to the obstacle. As shown in Figs.\ref{fig_discon}(c) and \ref{fig_discon}(d), the neurodynamics-based self-adaptive escape is able to achieve collision-free escape without member loss in the moving obstacle environment.

\section{Conclusion}
\label{sec:conclusion}
In this paper, a fish-inspired self-adaptive collective escape is developed for the swarm robots. The proposed method is motivated by the observation that a group of schooling fish can effectively complete escape tasks, despite limited communication with each other. A novel virtual forces approach is proposed to guide swarm robots to collision-free escape the threats through the dynamic landscape of neural activity. Moreover, a novel neurodynamics-based self-adaptive mechanism is proposed  to improve the escape performance of the swarm robots in changing environments. Simulation and experimental results demonstrate that the proposed approach allows for safe navigation and efficient self-adaptive cooperation among autonomous robots, enabling the successful completion of escape tasks in changing environments. 

 \bibliographystyle{IEEEtran}
\bibliography{IEEEabrv,reference_yuan_test}

 \vspace{-1.5cm}
    \begin{IEEEbiography}[{\includegraphics[width=1in,height=1.25in,clip,keepaspectratio]{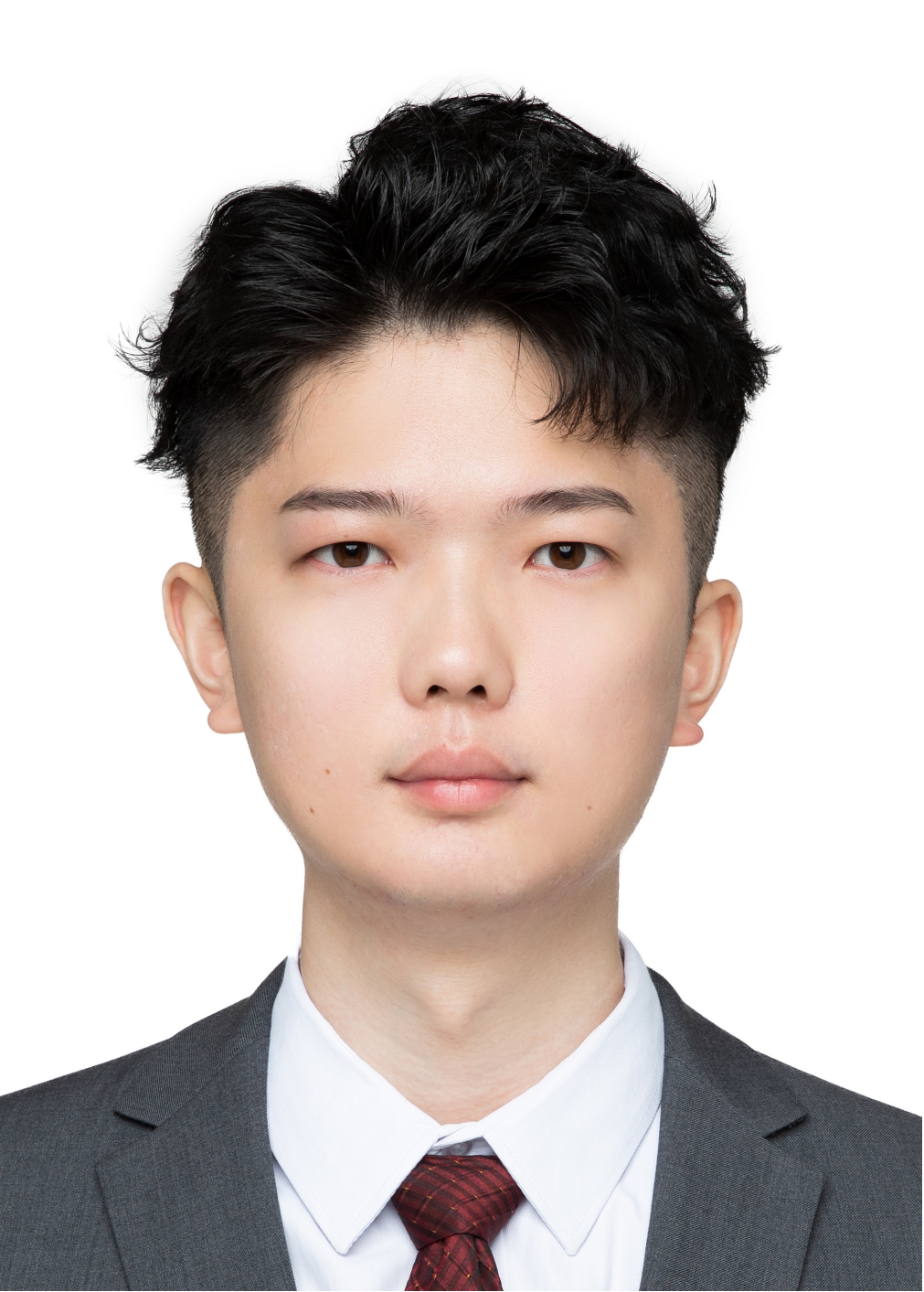}}]
    {Junfei Li} (Member, IEEE)  received the B.Eng. degree in communication engineering from Chongqing University of Posts and Telecommunications, Chongqing, China, in 2017, and the Ph.D. degree in engineering systems and computing from the University of Guelph, Ontario, Canada, in 2023. He is currently a Postdoctoral Research Fellow at the School of Engineering, University of Guelph, Ontario, Canada. His research interests include escape behaviors, search and rescue, and bio-inspired algorithms.


\end{IEEEbiography}

\vspace{-3cm}
\begin{IEEEbiography}[{\includegraphics[width=1in,height=1.25in,clip,keepaspectratio]{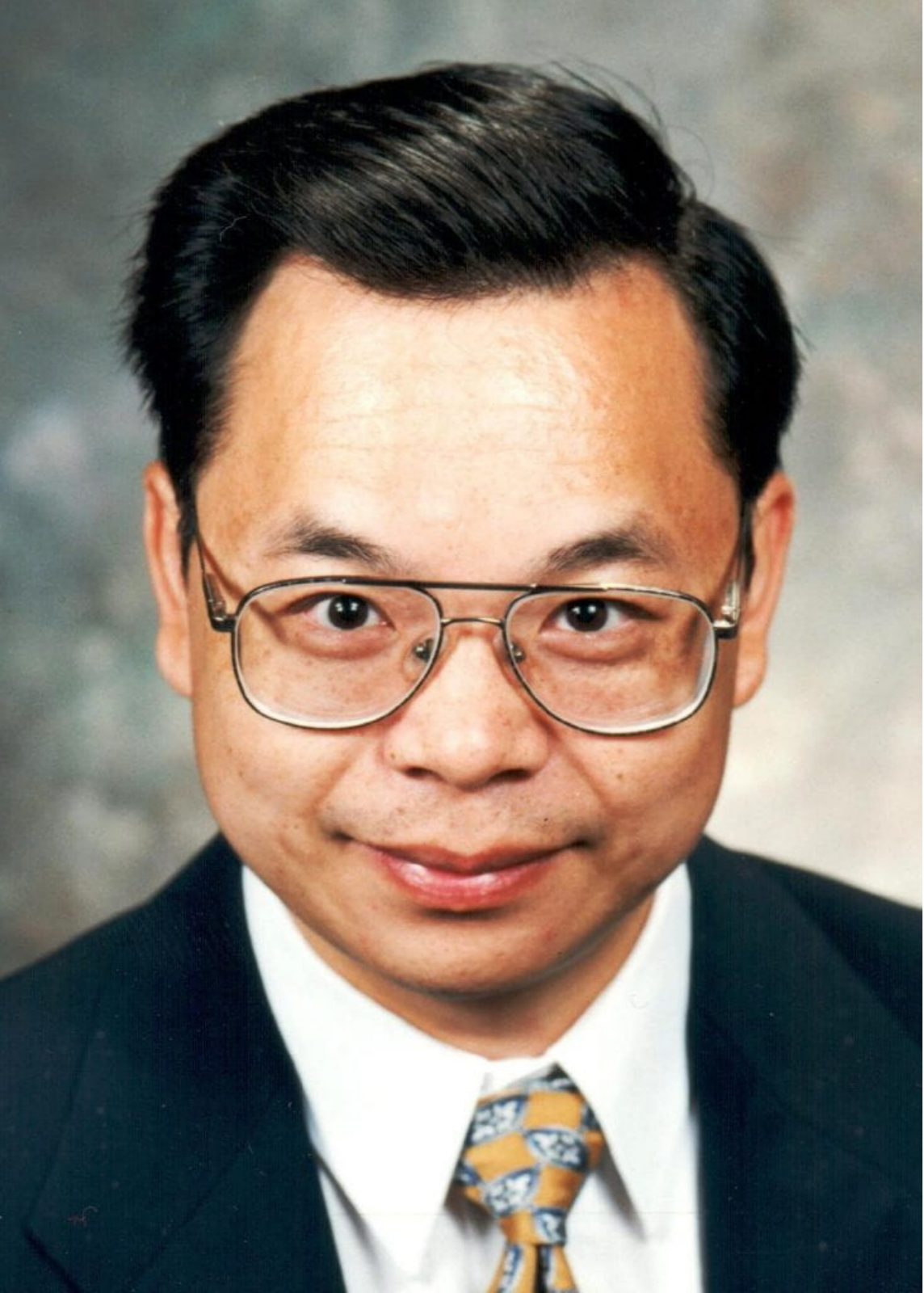}}]
{Simon X. Yang} (Senior Member, IEEE)  received the B.Sc. degree in engineering physics from Beijing University, Beijing, China, in 1987, the first of two M.Sc. degrees in biophysics from the Chinese Academy of Sciences, Beijing, in 1990, the second M.Sc. degree in electrical engineering from the University of Houston, Houston, TX, in 1996, and the Ph.D. degree in electrical and computer engineering from the University of Alberta, Edmonton, AB, Canada, in 1999.  He is currently a Professor and the Head of the Advanced Robotics and Intelligent Systems (ARIS) Laboratory at the University of Guelph, Guelph, ON, Canada. His research interests include robotics, intelligent systems, control systems, sensors and multi-sensor fusion, wireless sensor networks, intelligent communication, intelligent transportation, machine learning, fuzzy systems, and computational neuroscience. 

Prof. Yang he has been very active in professional activities. He serves as the Editor-in-Chief of Intelligence $\&$ Robotics, and International Journal of Robotics and Automation, and an Associate Editor of IEEE Transactions on Cybernetics, IEEE Transactions of Artificial Intelligence, and several other journals. He has involved in the organization of many international conferences.

\end{IEEEbiography}
\end{document}